 \pdfoutput=1

\documentclass[11pt]{article}

\usepackage[preprint]{acl}

\usepackage{svg}
\usepackage{times}
\usepackage{latexsym}
\usepackage{paralist}
\usepackage[utf8]{inputenc}
\usepackage[russian,english]{babel}
\usepackage[T1]{fontenc}
\usepackage{algorithm}
\usepackage{graphicx}
\usepackage{algpseudocode}
\usepackage{comment}
\usepackage{placeins}
\usepackage{subcaption}
\usepackage{multirow}
\usepackage{float}

\usepackage{microtype}

\usepackage{inconsolata}

\newcommand{\tool}{AcrosticSleuth}

\newcommand{\dataset}{AcrostID}

\title{AcrosticSleuth: Probabilistic Identification and Ranking of Acrostics in Multilingual Corpora}

\author{Aleksandr Fedchin \\
  Tufts University \\
  Medford, MA, USA \\
  \texttt{aleksandr.fedchin@tufts.edu} \\\And
  Isabel Cooperman \\
  University of Wisconsin Madison \\
  Madison, WI, USA \\
  \texttt{icooperman@wisc.edu  } \\\AND
  Pramit Chaudhuri \\
  University of Texas at Austin \\
  Austin, TX, USA \\
  \texttt{pramit.chaudhuri@austin.utexas.edu } \\\And
  Joseph P. Dexter \\
  Harvard University \\
  Cambridge, MA, USA \\
  \texttt{jdexter@fas.harvard.edu} \\} 

\newif\ifcomments
\commentsfalse 

\begin{document}
\maketitle
\begin{abstract}
For centuries, writers have hidden messages in their texts as acrostics, where initial letters of consecutive lines or paragraphs form meaningful words or phrases.
Scholars searching for acrostics manually can only focus on a few authors at a time and often favor qualitative arguments in discussing intentionally. 
We aim to put the study of acrostics on firmer statistical footing by presenting \tool{}, a first-of-its-kind tool that automatically identifies acrostics and ranks them by the probability that the sequence of characters does not occur by chance (and therefore may have been inserted intentionally).
Acrostics are rare, so we formalize the problem as a binary classification task in the presence of extreme class imbalance.
To evaluate \tool{}, we present the Acrostic Identification Dataset (\dataset{}), a collection of acrostics from the WikiSource online database. 
Despite the class imbalance, \tool{} achieves F1 scores of 0.39, 0.59, and 0.66 on French, English, and Russian subdomains of WikiSource, respectively.
We further demonstrate that \tool{} can identify previously unknown high-profile instances of wordplay, such as the acrostic spelling \texttt{ARSPOETICA} (``art of poetry") by Italian Humanist Albertino Mussato and English philosopher Thomas Hobbes' signature in the opening paragraphs of \emph{The Elements of Law}.
\end{abstract}

\section{Introduction}

If you put together the initial letters of the 14 opening paragraphs of Thomas Hobbes' \emph{The Elements of Law}, you will discover that they spell \texttt{THOMAS[OF]HOBBES}.
Such hidden messages, where initial letters of lines or paragraphs spell a meaningful word or phrase, are called \emph{acrostics}. 
Acrostics are easy to find if you know where to look -- some authors even draw attention to them -- but can otherwise be difficult to notice. 
For example, we are the first, to our knowledge, to identify the Hobbes acrostic, despite its appearance at the beginning of an important, well-studied text by a famous thinker.
The subtle or playful nature of acrostics has kept the literary device in regular if infrequent use throughout the centuries. 
Most recently, Russian dissidents have inserted anti-government messages as acrostics in mainstream publications.\footnotemark
\footnotetext{
To cite two examples, 
politically persecuted film director and LGBT activist Kirill \citet{serebrennikov} encoded a message in his final speech to the court that spells \foreignlanguage{russian}{\texttt{НИОЧЕМНЕЖАЛЕЮСОЧУВСТВУЮВАМ}} (``I have no regrets. I am sorry for you") and scholar Ilya Lemeshkin \citep{zona2023publication} published a paper in a government-funded journal, with an acrostic \foreignlanguage{russian}{\texttt{СДОХНИПУТЛЕРНЕТВОЙНЕИЛ}} (``Die Putler. No to war. I.L.").
}

In contrast to these unambiguous examples, scholars have also argued for the intentionality of much shorter acrostics, such as the supposed acrostic \texttt{MARS} in the middle of Vergil's {\it Aeneid} \citep{marsacrostic}.
Critics have seen the use of two regular Latin terms for war within the passage (Martem, bellum) as validating the acrostic, but without any attention to the probability of the four-letter sequence.
We further discuss related work in Section~\ref{sec:background}.
To our knowledge, no study exists that takes a systematic, quantitative approach to both the identification and analysis of acrostics across multiple languages.

\begin{figure}[htb]
\includegraphics[width=\linewidth]{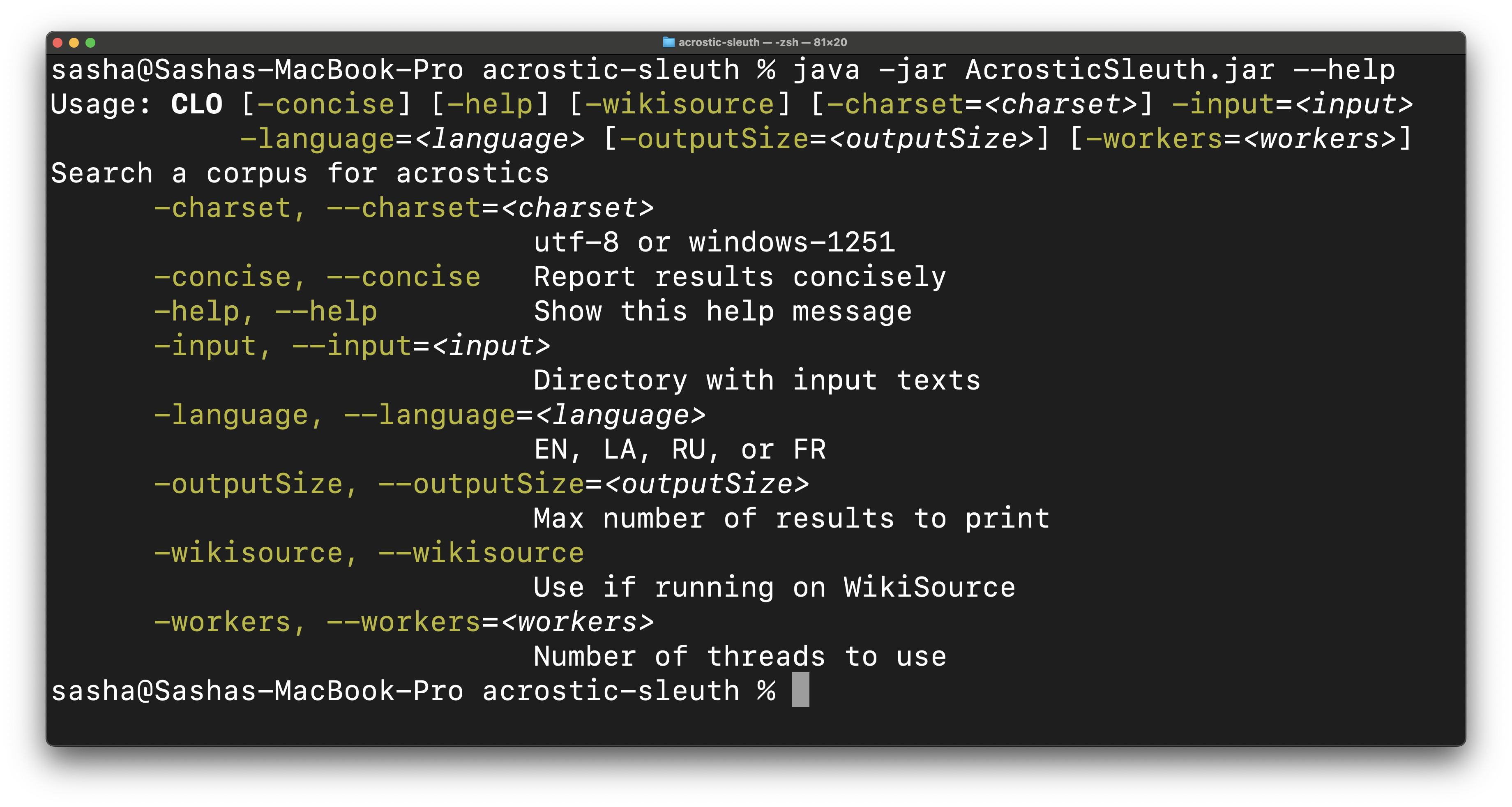}
\caption{Screenshot showing \tool{}'s help message and options.}
\label{fig:screenshot}
\end{figure}

In this paper, we introduce \tool{}, a tool that can process large corpora of texts, identify hypothetical acrostics, and rank them by the probability that the sequence of initial characters does not occur by chance (and therefore may have been inserted intentionally by the author).
\tool{} is a command line tool (see the screenshot in Figure~\ref{fig:screenshot}) available on GitHub under the MIT license.\footnote{\url{https://github.com/acrostics/acrostic-sleuth}}
From a statistical perspective, the acrostic identification problem presents a challenge in the form of extreme class imbalance -- acrostics are very rare. 
In Section~\ref{sec:methods}, we discuss the steps we take to identify and rank acrostics, as well as the implementation details that allow the search to be efficient.

To evaluate \tool{}, we create Acrostic Identification Dataset (\dataset{}), a collection of labelled acrostics from the English, French, and Russian subdomains of the WikiSource database of texts. 
The dataset is available under the MIT license and includes all acrostics that have been explicitly referred to or formatted as such on WikiSource.\footnote{\url{https://github.com/acrostics/acrostic-identification-dataset}}
We show that \tool{} successfully identifies acrostic poems and achieves F1 scores of 0.39, 0.59, and 0.66 on French, English, and Russian corpora, respectively.
In Section~\ref{sec:evaluation}, we present these results, provide a comparison of the tool's performance across languages, and discuss acrostics found by \tool{} that have not previously been recognized. In Section~\ref{sec:discussion}, we discuss the implications that this work has for the study of wordplay and outline directions for future work.

In summary, the main contributions of this paper are as follows:
\begin{compactitem}
\item We formalize the problem of identifying acrostics as a binary classification task in the presence of extreme class imbalance.
\item We present \tool{}, an efficient and publicly licensed tool for finding and ranking acrostics.
\item We present \dataset{}, a multilingual dataset of acrostics that can be used for the study of this form of wordplay.
\item We evaluate \tool{} on \dataset{} and show that the tool successfully identifies real-world instances of acrostics.
\end{compactitem}

\section{Background}
\label{sec:background}

Attempts have been made in the past to employ quantitative analysis in discussing intentionality of acrostics, including in Shakespeare~\citep{eckler} and Horace~\citep{morgan1993}, among others. 
Such studies typically base their arguments on calculating the probability of encountering an acrostic by pure chance \citep{morgan1993} (as opposed to ranking them).
The problem with this approach, as noted by Matthew \citet{robinson}, is that, while the probability of any given acrostic is indeed very low, one is nevertheless almost guaranteed to stumble upon some accidental acrostic in a text of any considerable length.
The lottery offers a good analogy: the chances of winning are abysmally low for any one person, but someone still takes home the jackpot. 

Our work falls under the broad category of automated analysis of wordplay and puzzles, with \tool{} being similar to, e.g., crossword solving tools~\cite{kulshreshtha2022down}. On the other end of the spectrum, there is a substantial body of work on development of language models that can compose acrostic poems \citep{acrostic-generation, acrostic-generation2}, paraphrase existing texts to introduce acrostics \citep{acrostic-paraphrase}, generate anagrams \citep{anaram-generate}, or synthesize other kinds of wordplay \citep{chinese-poetry-generate}. The wide availability of such tools and the drive for creative language encodings intended to avoid censorship~\citep{censorship} in online communication suggest that acrostics may become even more widespread in the future. 

\section{Methods}
\label{sec:methods}

\begin{figure*}[htb]
\includegraphics[width=\linewidth]{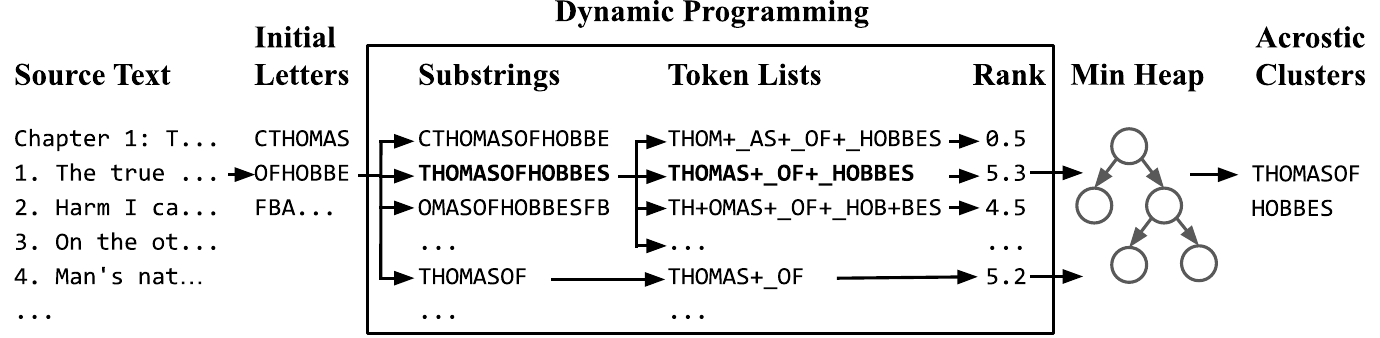}
\caption{\tool{}'s Workflow.}
\label{fig:workflow}
\end{figure*}

In this section, we outline our approach for enumerating and ranking candidate acrostics. 
We can formally define the problem as follows: given a sequence of characters with which lines in a text begin, rank all possible subsequences based on the probability that they come from natural speech and have not been selected randomly from the distribution of initial letters. 
Our hypothesis is that this probability should reflect intentionality: the higher the probability, the more likely it is that the corresponding characters have been deliberately made to form meaningful words or phrases by the author.
In Section~\ref{sec:methods:overview} we discuss what it means to compare and rank a pair of character sequences from a theoretical perspective.
In Section~\ref{sec:methods:lm} we justify our choice of the language model.
Finally, in Section~\ref{sec:methods:implementation} we discuss implementation details that allow us to perform the ranking efficiently.

\subsection{Overview}
\label{sec:methods:overview}
Consider the binary classification problem of labeling a sequence of characters as ``acrostic'' or ``not an acrostic''.
Note that this problem suffers from extreme class imbalance -- most sequences of characters will not be acrostics.
Since we do not know the a priori probability $P(a)$ of encountering an acrostic, we cannot directly compute $P(a|s)$ or the probability that a given sequence of characters $s$ is an acrostic.
In order to rank candidate acrostics, however, it is enough to estimate $\frac{P(a|s_1)}{P(a|s_2)}$, the ratio of two such probabilities for two different strings $s_1$ and $s_2$.
By Bayes theorem, this ratio is equal to $\frac{P(s_1|a)P(a)P(s_2)}{P(s_1)P(s_2|a)P(a)}=\frac{P(s_1|a)P(s_2)}{P(s_1)P(s_2|a)}=\frac{P(s_1|a)(P(s_2|a)P(a) + P(s_2|\neg a)P(\neg a))}{P(s_2|a)(P(s_1|a)P(a) + P(s_1|\neg a)P(\neg a))}$.
We now posit that $P(a)$, the a priori probability of encountering an acrostic, must be very small.
As $P(a)$ approaches zero, the ratio we need to estimate approaches $\frac{P(s_1|a)P(s_2|\neg a)}{P(s_2|a)P(s_1|\neg a)}$.
In other words, computing $\frac{P(s|a)}{P(s|\neg a)}$ for every string $s$ gives us a metric by which we can rank all candidate acrostics.
We will refer to this ratio as the \emph{rank} of an acrostic throughout the paper.

Of the two probabilities involved in computing the rank, estimating $P(s|\neg a)$ for some $s$ is trivial -- it is the conditional probability of first letters in each line forming the sequence of characters $s$ under the assumption that the poem contains no acrostics.
When there are no acrostics, each character in $s$ is drawn independently at random from the overall distribution of characters with which English (or French, Russian, etc.) words begin, so $P(s|\neg a)$ is a product of such probabilities for individual characters. On the other hand, $P(s|\neg a)$, the probability of encountering a sequence of characters in an acrostic, is more difficult to estimate. 
In this study, we assume that acrostics are similar to regular text, which allows us to use pre-trained language models.
We discuss our choice of the language model in the next section.

\subsection{SentencePiece Unigram Language Model}
\label{sec:methods:lm}

\tool{} relies on unigram language models produced by SentencePiece~\citep{sentencepiece} to estimate the probability that an acrostic spells some given sequence of characters. 
SentencePiece is an unsupervised text tokenizer, where subword level tokens are chosen to fit the vocabulary size specified by the user.
In selecting SentencePiece, we made the following considerations:
\begin{compactitem}
\item \textbf{No Supervision.} SentencePiece is fully unsupervised, making it easy to adapt \tool{} to multiple languages.
\item \textbf{Subword Tokens.} SentencePiece uses subword tokens, meaning that it can deal with out-of-vocabulary words by decomposing them into syllables or suffixes. 
This feature is particularly useful in the context of acrostics, since acrostics often refer to people, whose names might not appear anywhere else in the corpus.
If we were to use a word-level tokenizer, our tool would miss out on many fascinating examples, some of which are discussed below.
\item \textbf{Speed.} Estimating probabilities with a unigram language model involves very little computational effort, allowing \tool{} to process large amounts of text quickly. 
Unigram models also allow dynamic programming, as we discuss in the next section.
\end{compactitem}

We found that \tool{} achieves best performance when using language models with large numbers of tokens.
The Appendix contains further discussion of how the model size affects \tool{}'s performance.

As an illustration of our reasoning, consider the \textit{lorem ipsum} placeholder text often used in publishing.
With a SentencePiece unigram model as a backend, \tool{} would assign such text a much higher rank than it would to a text you might get by drawing from the Latin alphabet at random.
We hypothesize that in most cases, this is enough to distinguish an acrostic from random noise. 

\subsection{Implementation}
\label{sec:methods:implementation}

Figure~\ref{fig:workflow} uses Hobbes' \emph{The Elements of Law} as an example to illustrate \tool{}'s workflow. 
\tool{} first converts the source text into a string of initial letters.
This step involves minimal language-specific preprocessing such as removing all non-alphabetic characters, etc. 
Next, \tool{} considers every possible substring of initial letters up to some fixed length and, for each substring, every possible way to tokenize it. 
For each resulting list of tokens, \tool{} computes the rank as discussed in Section~\ref{sec:methods:overview}.
This step is done in a dynamic programming fashion, where we process substrings that end earlier in the text first and store the highest ranking tokenization of previously encountered substrings. 
Note that the use of a unigram model significantly simplifies dynamic programming setup, since the probability assigned to the next token is independent from the previous ones.
We further boost the performance by supporting multithreading (each thread processes its own text) and maintaining a cache of commonly occurring substrings.

\tool{} uses a min-heap data structure to keep track of the highest ranking candidate acrostics it has encountered so far. 
The size of the heap is fixed and can be specified by the user.
When reporting the results, \tool{} aggregates candidate acrostics that overlap into one result and uses the highest-ranked candidate acrostic to rank the whole cluster. 
For example, in Figure~\ref{fig:workflow} both \texttt{THOMASOF} and \texttt{THOMASOFHOBBES} end up in the min-heap as high-ranking candidate acrostics, but \tool{} reports them as one result cluster because they overlap.

\section{Results and Evaluation}
\label{sec:evaluation}

In evaluating \tool{}, we aimed to answer the following research questions:
\begin{compactitem}
\item \textbf{RQ1:} How successfully does the tool identify already known acrostics?
\item \textbf{RQ2:} Does the tool uncover previously unknown acrostics?
\item \textbf{RQ3:} How does the tool's performance differ across languages?
\end{compactitem}

\begin{figure}[htb]
\includegraphics[width=\linewidth]{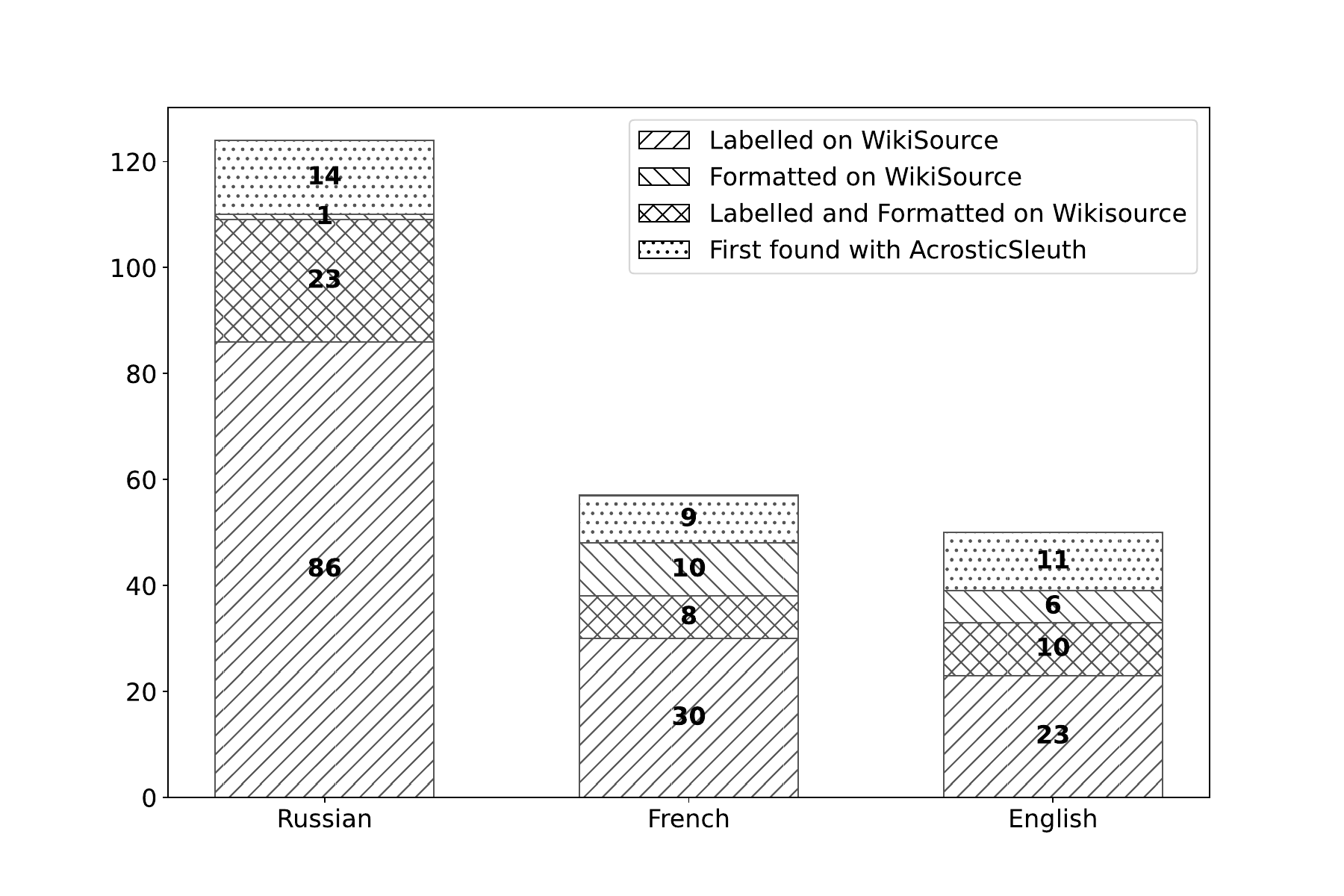}
\caption{The method by which acrostics in \dataset{} have been identified.}
\label{fig:results}
\end{figure}

\begin{table*}[htb]
\small{}
\begin{tabular}{| r | r | r |}
\hline
Acrostic						& 	WikiSource Page	\\ \hline \hline
\texttt{TOJOSEPHKNIGHT}					&	Page:Notes by the Way.djvu/61 \\ \hline
\texttt{IESUCHRISTSONNEOFGODTHESAVIOR}	&	Page:Whole prophecies of Scotland, England, Ireland, France \& Denmark.pdf/46 \\\hline 
\texttt{PRINCECHARLIE}					&	Page:Carroll - Three Sunsets.djvu/83 \\\hline
\texttt{CORNELIABASSET}					&	Ben King's Verse/Asphodel \\\hline
\texttt{KATHLEENBRUCE}					&	Page:Clouds without Water (Crowley, 1909).djvu/24 \\\hline
\texttt{AMAZING}						&	Page:Amazing Stories Volume 17 Number 06.djvu/6 \\\hline
\texttt{PERHAPS}						&	Page:Love's trilogy.djvu/79 \\\hline
\texttt{ALICEPLEASANCELIDDELL}				&	Page:Complete Works of Lewis Carroll.djvu/292\\\hline 
\texttt{THOMAS[OF]HOBBES}				&	The Elements of Law/Part I/Chapter 1\\\hline
\texttt{MARYSTOKES}					&	Page:Notes and Queries - Series 12 - Volume 4.djvu/257\\ \hline
\multirow[t]{2}{*}{\texttt{SURVIVAL}}						&	United States Army Field Manual 7-93 Long-Range Surveillance Unit Operations/\\ & Appendix F \\\hline
\end{tabular}
\caption{English acrostics that have not been labeled or formatted as such on WikiSource as of April 20th 2024}
\label{tab:english-results}
\end{table*}

\subsection{Acrostic Identification Dataset}
\label{sec:evaluation:dataset}

To evaluate \tool{}'s performance, we created Acrostic Identification Dataset (\dataset{}), which is comprised of acrostics found on WikiSource, a Wikipedia-affiliated online library of literature, parliamentary proceedings, and other source texts.
In choosing WikiSource as the base for the dataset, we considered the following criteria:
\begin{compactitem}
\item \textbf{Availability and reproducibility}. WikiSource is publicly licensed and hosts timestamped copies of itself, facilitating reproduction.
\item \textbf{Multilingual coverage}. WikiSource contains texts written in multiple languages, allowing cross-lingual comparisons. 
Specifically, we analyzed English, French, and Russian subdomains of WikiSource.
\item \textbf{Applicability}. A large portion of WikiSource consists of poetry, which historically is the genre best known for acrostics.
\item \textbf{Annotation.} WikiSource is partially annotated: multiple poems and texts are explicitly labeled or formatted as acrostics, which allows measuring the recall of an acrostic-identification tool. 
\end{compactitem}

To obtain a set of ``true" acrostics on which we could evaluate our tool, we performed the following two tasks.
First, we manually inspected all uses of the word ``acrostic" on WikiSource (``\foreignlanguage{russian}{акростих}" in Russian, ``acrostiche" in French).
In cases when the word referred to specific lines, we marked down these lines as acrostics. 
This method allowed us to identify 33 acrostics in English WikiSource, 109 in Russian, and 38 in French.

We have also identified acrostics based on formatting: initial letters of an acrostic are often highlighted in bold or in red or are rotated by 90 degrees.
We looked at all sequences of 5 or more consecutive lines where initial letters are specially formatted, and identified cases where the initial letters form a word/words in the source language.\footnote{Over 90\% of labelled acrostics on WikiSource are at least 5 letters long, and inspecting shorter sequences would be prohibitively time consuming}
Finally, we manually inspected the corresponding WikiSource pages to confirm that the formatting is not accidental.
This method allowed us to identify a further 6 English acrostics, as well as 1 Russian and 10 French. 

Figure~\ref{fig:results} summarizes these counts for each language and also shows the number of new acrostics that \tool{} identifies. 
We include these new acrostics in the dataset but mark them separately so as not to count them towards recall when evaluating the tool.
We discuss these new discoveries in greater detail in Section~\ref{sec:evaluation:experiments}

When reporting the number of acrostics above, we group some acrostics together to count them as one.
Specifically, we do this when any of the following is true: (i) a single acrostic is split between multiple WikiSource pages, (ii) the same acrostic is reproduced multiple times on several WikiSource pages, or (iii) two separate acrostics are within 10 lines of each other on the same WikiSource page. 
Our reason for grouping these together is that an acrostic identification tool should not be rewarded or penalized more for finding or failing to find an acrostic reproduced multiple times, split into multiple pages, or made easier to discover due to being collocated with another acrostic. 

\subsection{Experiments}
\label{sec:evaluation:experiments}

We performed all experiments on an M3 Mac with 48 GB RAM and 16 CPU cores. 
All experiments can be reproduced in under an hour using comparable resources.

\begin{figure}[htb]
\begin{subfigure}{0.96\linewidth}
\includegraphics[width=\linewidth]{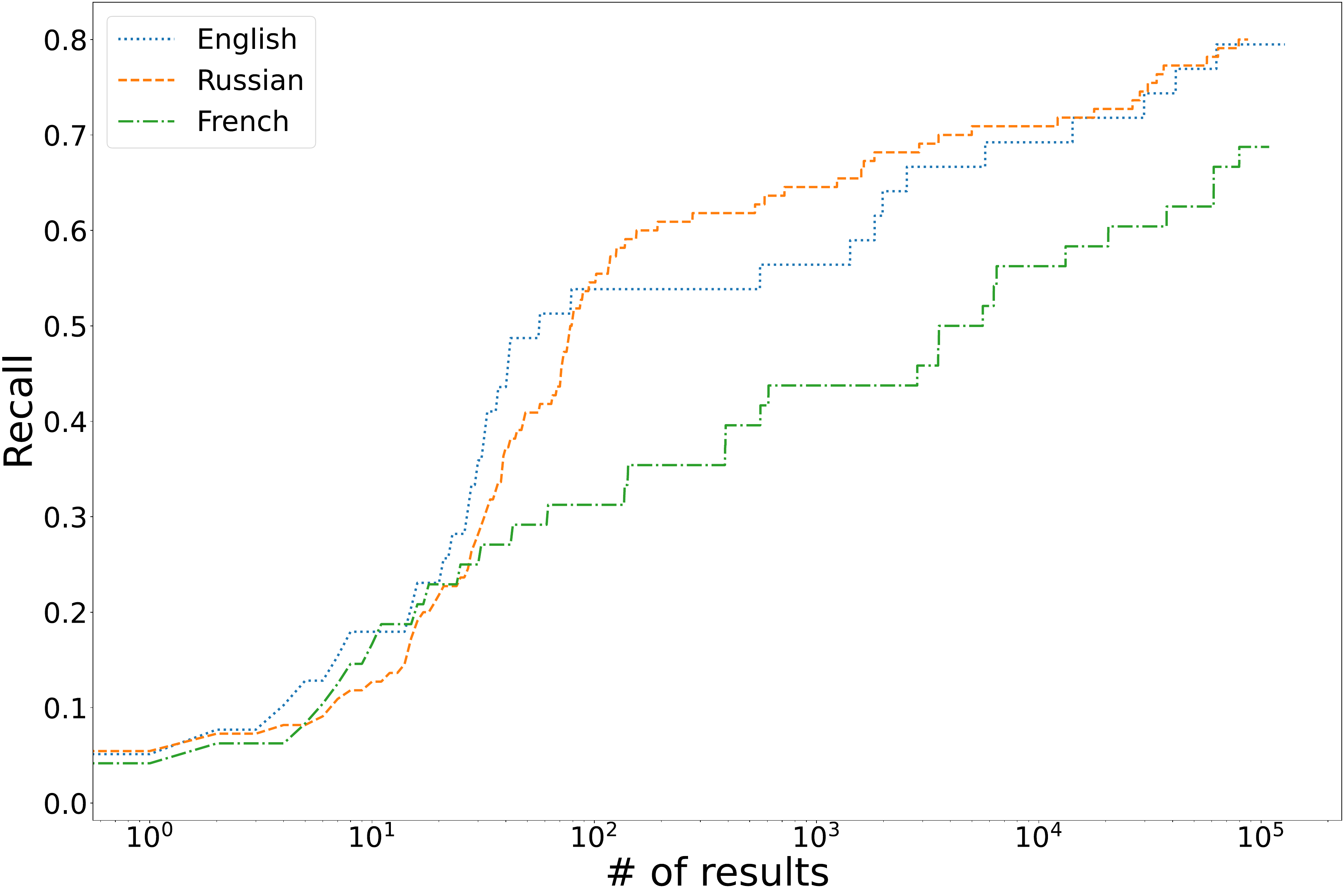}
\caption{Recall.}
\label{fig:evaluation:recall}
\end{subfigure}
\begin{subfigure}{0.96\linewidth}
\includegraphics[width=\linewidth]{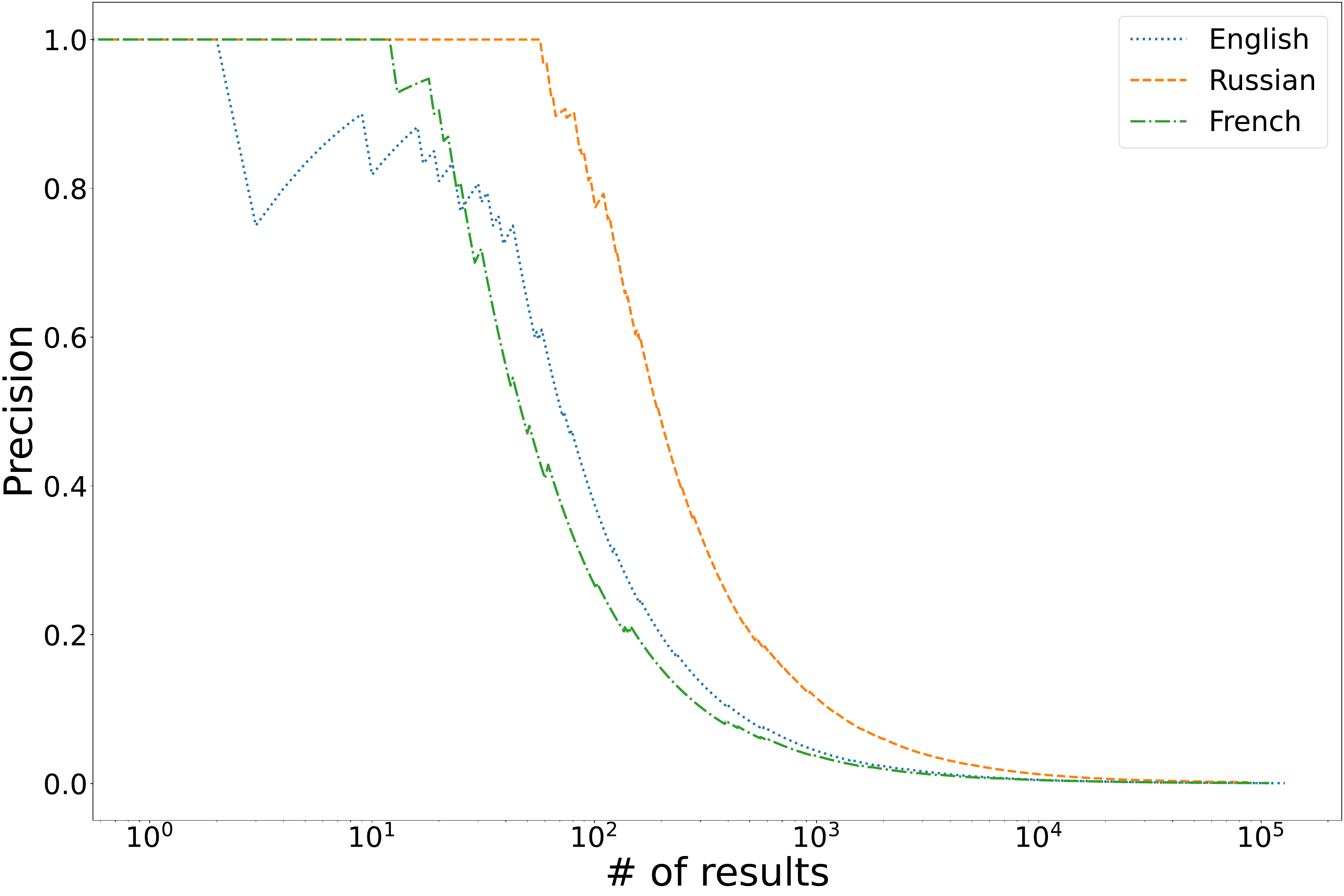}
\caption{Precision.}
\label{fig:evaluation:precision}
\end{subfigure}
\begin{subfigure}{\linewidth}
\includegraphics[width=0.96\linewidth]{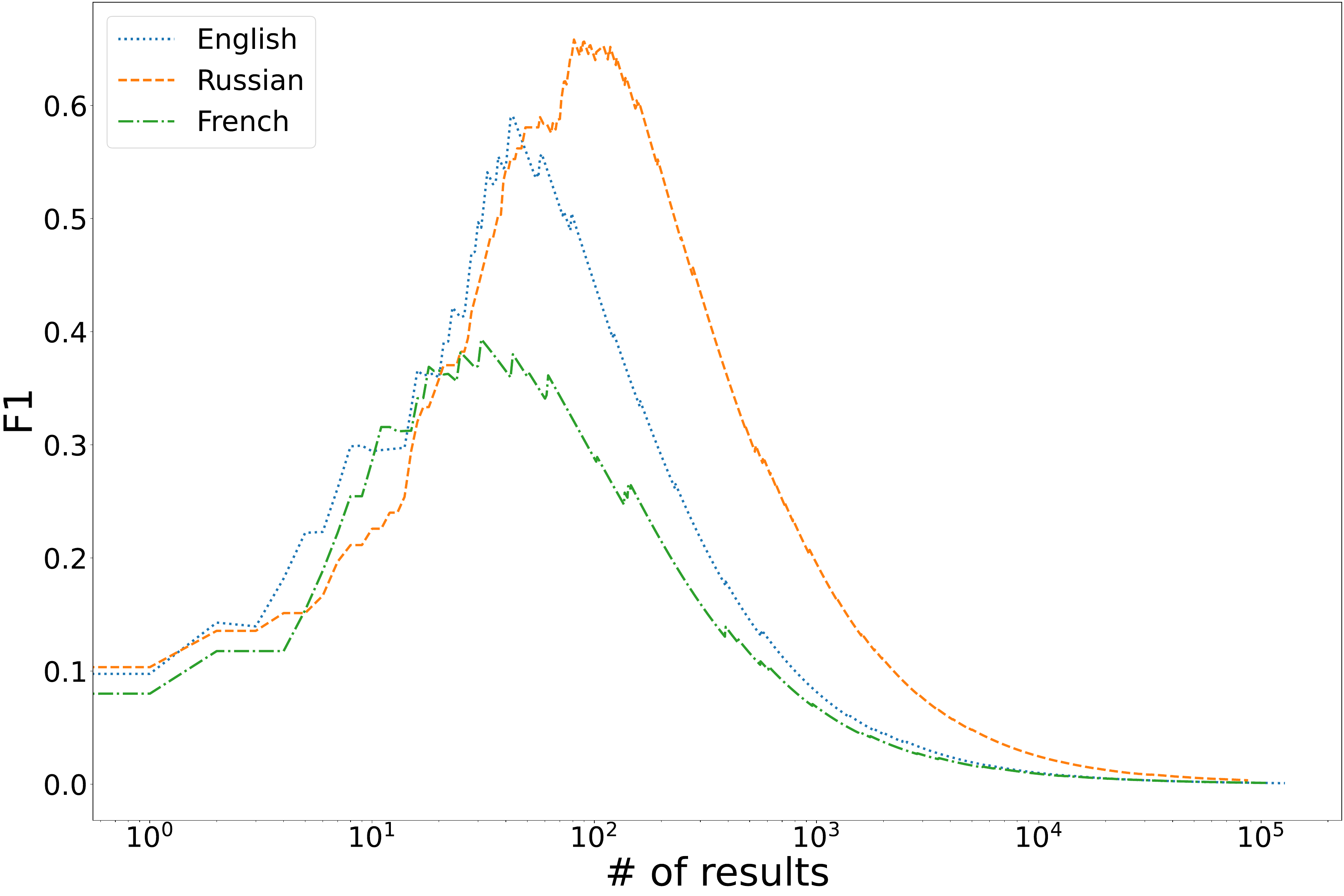}
\caption{F1 Score.}
\label{fig:evaluation:f1}
\end{subfigure}
\caption{Performance of the acrostic finding tool.}
\label{fig:evaluation}
\end{figure}

Figures~\ref{fig:evaluation:recall},~\ref{fig:evaluation:precision},~and~\ref{fig:evaluation:f1} summarize the recall, precision, and f1-score, respectively, that \tool{} achieves on the \dataset{}. 
On all subplots, the y-axis shows the tool's performance, and the x-axis (logarithmic scale) indicates the number of first-ranking results for which the corresponding metric is calculated.
Despite the extreme class imbalance, \tool{} achieves the top F1 scores of 0.39, 0.59, and 0.66 on French, English, and Russian corpora, respectively. 
With \tool{}'s help, a researcher can quickly identify the majority of acrostics in any given corpus.
We believe that this data positively answers \textbf{RQ1}.

Figure~\ref{fig:evaluation} clearly shows \tool{} achieves the best performance on the Russian corpus and the worst on the French.
We believe this difference to be primarily determined by the datasets themselves. 
In particular, Russian WikiSource contains a large number of 17th and 18th century acrostics, which tend to span dozens of lines and are thus easier to identify.
By contrast, many acrsotics on French WikiSource are split into multiple pages and are otherwise more difficult to identify due to formatting issues.
Therefore, to answer \textbf{RQ3}, we believe that \tool{} can easily be used with multiple languages but its success depends, at least in part, on the nature of the data.

When calculating recall, which is the main metric by which we believe the tool should be evaluated, we only considered as true acrostics those instances that we identified manually as described in Section~\ref{sec:evaluation:dataset}. 
When calculating precision, however, we also took into account acrostics that the tool identifies that are not labelled or formatted as such on WikiSource.
To do so, we manually inspected the top 1000 results \tool{} returns for each language and identified those we believe to be acrostics beyond any doubt.
Figure~\ref{fig:results} shows the number of all newly found acrostics for each language, and
Table~\ref{tab:english-results} lists all those found on the English subdomain of WikiSource. 
While we do not believe the intentionality of acrostics in Table~\ref{tab:english-results} could be a point of contention, we should report that the F1 score drops to 0.36, 0.48, 0.57 on French, English, and Russian, respectively, if we do not count these instances towards precision.
Note that some of the acrostics in Table~\ref{tab:english-results} have been identified before, such as that by Lewis Carroll, although the corresponding page on WikiSource contains no reference to an acrostic.
Other acrostics, however, are new discoveries, such as the \texttt{THOMAS[OF]HOBBES} example discussed in the Introduction.
These results positively answer \textbf{RQ2}.

\section{Discussion and Future Work}
\label{sec:discussion}

Throughout this paper, we discuss the acrostics that our tool identifies only insofar as they are relevant for the tool's evaluation.
A direction for future qualitative research is the analysis of acrostics and similar forms of wordplay in their own right.

One property shared by the majority of the acrostics we find is that they appear at the very beginning of their respective texts. 
This tendency is not surprising---readers are more likely to look for acrostics in the first lines of a poem---but it does highlight the idiosyncrasy of examples such as the \texttt{MARS} acrostic we mention in the introduction, which appears in the middle of the \textit{Aeneid}.
We have run \tool{} on Musisque Deoque and Poeti D'Italia, two databases of Latin poetry, and found that the tool ranked this acrostic very low, primarily because it is short.
Our point here is not to argue for or against intentionality of this specific instance but to emphasize the complementary nature of qualitative and quantitative analysis.
Despite ranking the much-discussed \texttt{MARS} acrostic low, \tool{} uncovered some previously unknown acrostics, such as the \texttt{ARSPOETICA} poem attributed to Albertino Mussato (Biblioteca Marciana XIV.223, edited by Padrin 1887).
The poem laments that Italy of Mussato's time is no longer safe for poets, and the acrostic identifies the female subject of the opening sentence, suggesting that she (the art of poetry) is not at ease.

Latin literature presents an interesting case study for other reasons. 
Roman authors composed not only regular acrostics but also telestics (formed by combining the final letters of each line), mesostics (formed by every n-th letter), diagonal acrostics, etc.  
Adapting \tool{} to rank all such alternative forms of wordplay together is a non trivial task that we plan to explore in the future. 

\section{Conclusion}

This paper presents \tool{}, a tool for identifying and ranking acrostics in large corpora of texts. 
We show that our implementation can not only identify well-known acrostics, but also uncovers high-profile instances that have not been discussed before, such as Hobbes' \texttt{THOMAS[OF]HOBBES} or Mussato's \texttt{ARSPOETICA}.

\section{Ethical Considerations}
In the introduction, we mention that acrostics have been used by dissidents to insert anti-war messages in mainstream media. 
In theory, one can imagine a malicious actor using a tool such as \tool{} to screen incoming publications for ``undesirable'' acrostics. 
In practice, however, \tool{} can only identify one specific kind of acrostics--those formed by initial letters of each line or paragraph. 
Other kinds of acrostics, such as those formed by initial letters of each word or sentence, would remain undetected. 
Moreover, even if the tool's functionality were further extended to cover these cases as well, one could always come up with a new way to encode a hidden message into the text. 
The point of acrostics by Lemeshkin and Serebrennikov, for instance, is precisely that there is no way to silence or prevent such artistic expressions of opinion, regardless of how much effort one spends on censorship.
These types of examples are not hidden in the sense of wishing to evade all notice either---they are calculated to provoke or amuse, with Lemeshkin publicly revealing his acrostic right after publication.
Hence, in the unlikely event that \tool{} were used for censorship purposes, we believe that the effort would prove to be futile.

\section*{Acknowledgements}

We thank Kinch Hoekstra and Johann Sommerville for their help in confirming that the Hobbes' acrostic has not been discovered before and their insights into its significance.

\bibliography{references}

\newpage

\appendix

\section*{Appendix}

\subsection*{Effect of Model's Size on Performance}
\label{sec:apdx:model}

\begin{figure}[htb]
\begin{subfigure}{\linewidth}
\includegraphics[scale=0.12]{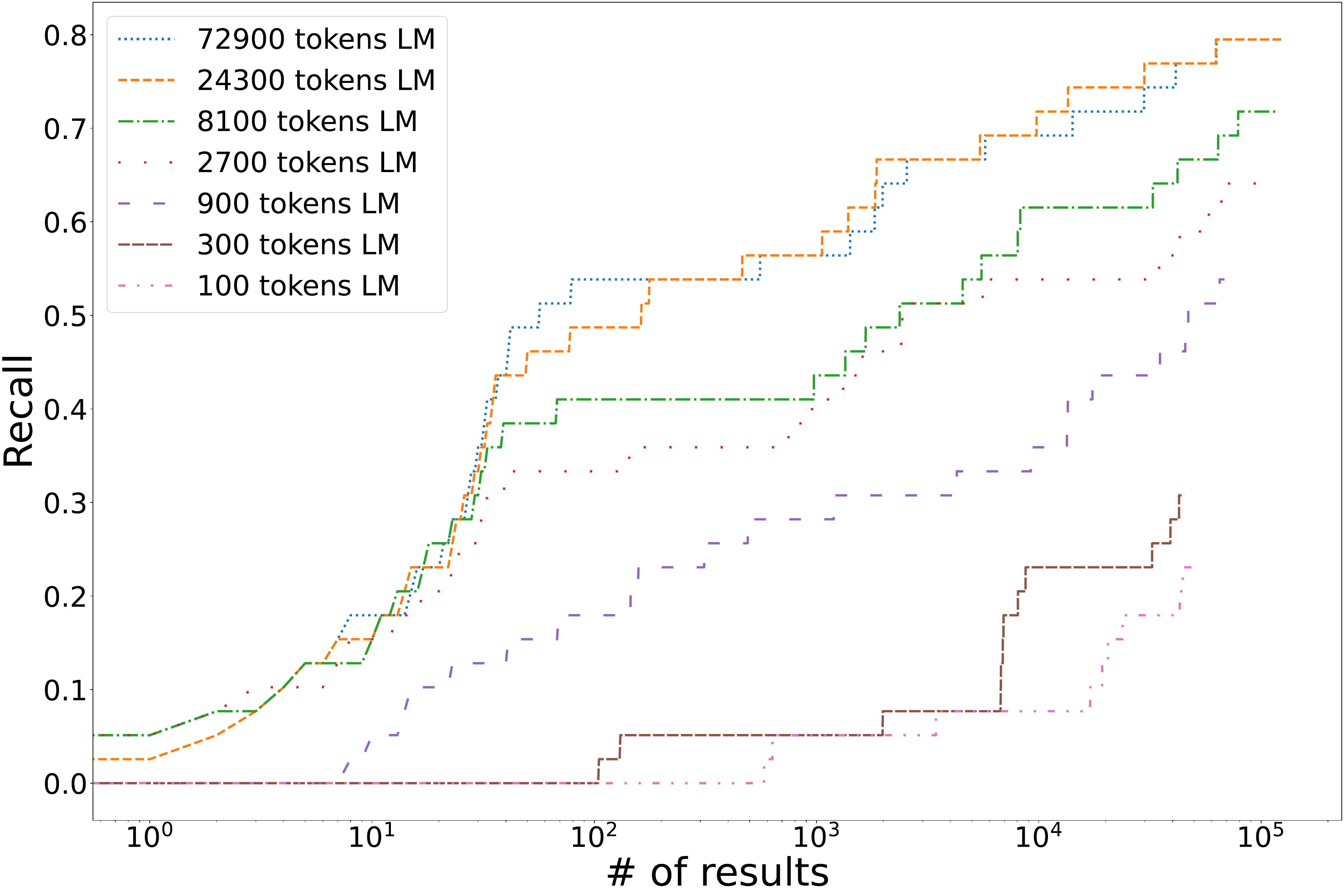}
\caption{English.}
\label{fig:evaluation:english}
\end{subfigure}
\begin{subfigure}{\linewidth}
\includegraphics[scale=0.12]{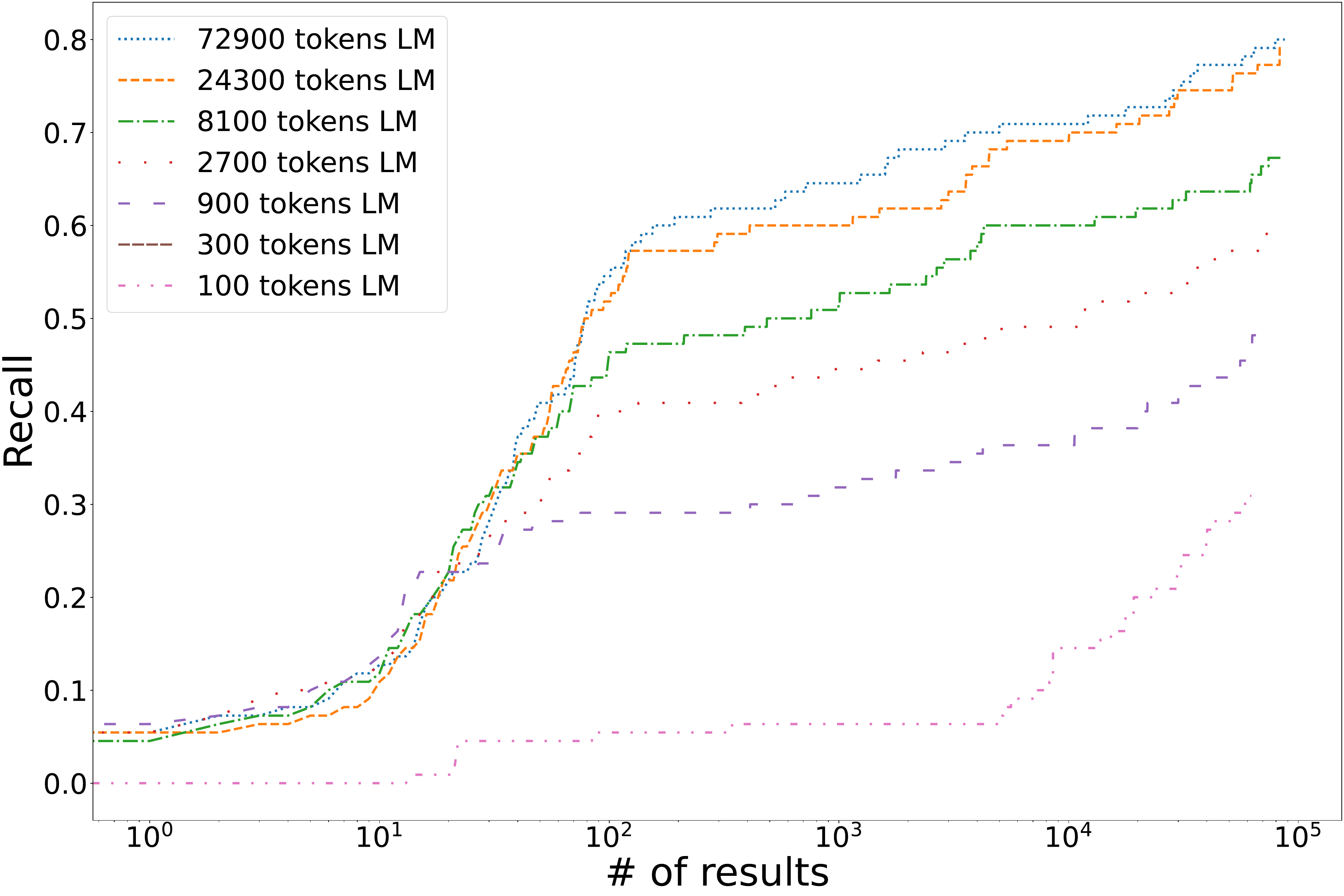}
\caption{Russian}
\label{apds:fig:evaluation:russian}
\end{subfigure}
\begin{subfigure}{\linewidth}
\includegraphics[scale=0.12]{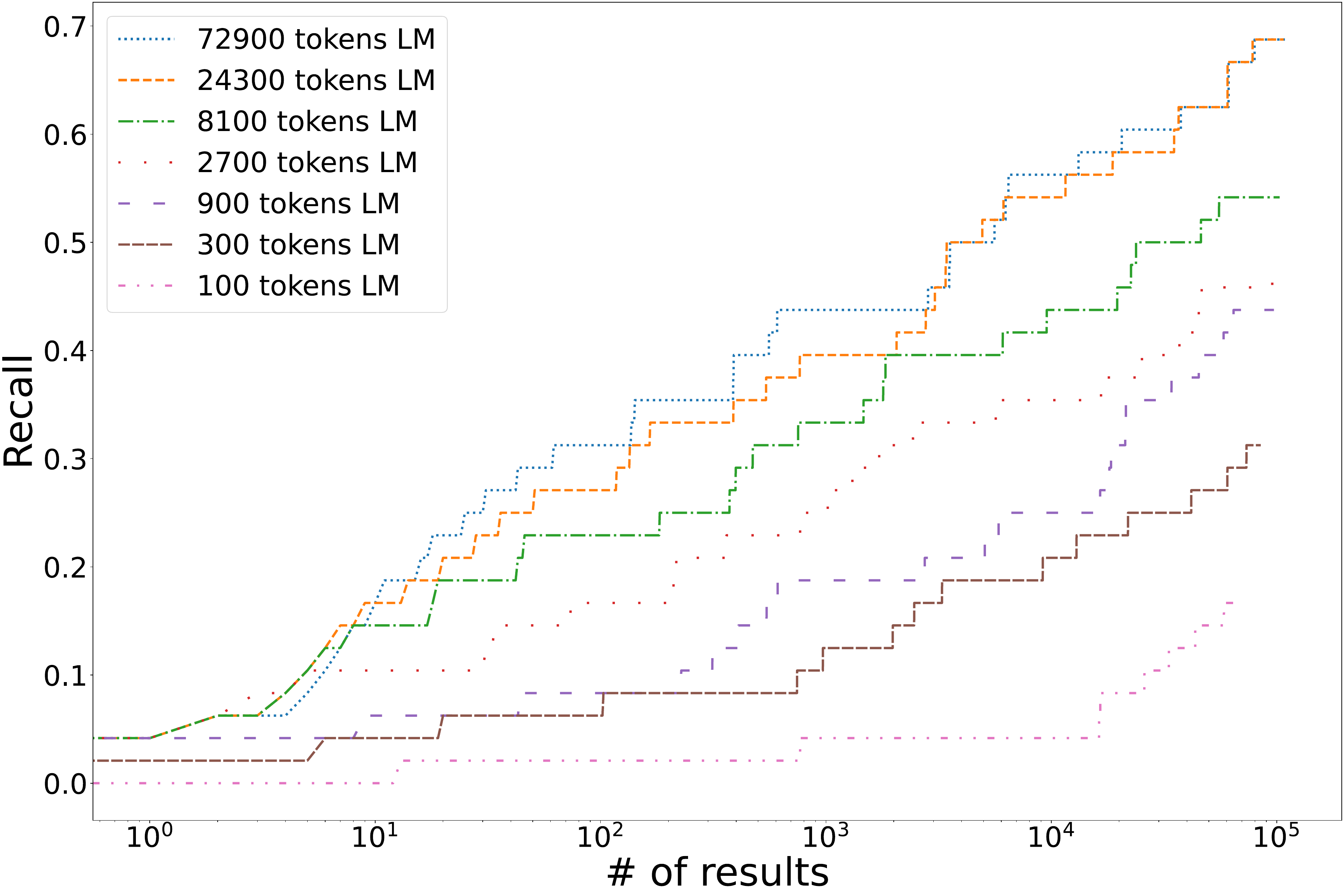}
\caption{French}
\label{apds:fig:evaluation:french}
\end{subfigure}
\caption{Effect of language model size on \tool{}'s recall for English, Russian, and French corpora.}
\label{apdx:fig:evaluation}
\end{figure}

In Section~\ref{sec:methods:lm}, we write that \tool{} reaches peak performance when using SentencePiece language models with the largest number of tokens. 
Figure~\ref{apdx:fig:evaluation} demonstrates this point further by showing the recall that \tool{} achieves for different languages with models of different sizes. 
Note that the performance of the largest two models (72900 and 24300 tokens respectively) is very similar, suggesting that further increasing the model's size would yield diminishing returns. 

\end{document}